\documentclass{article}

\PassOptionsToPackage{numbers, compress}{natbib}

\usepackage[final]{NFFL2021}

\usepackage[utf8]{inputenc} 
\usepackage[T1]{fontenc}    
\usepackage{hyperref}       
\usepackage{url}            
\usepackage{booktabs}       
\usepackage{amsfonts}       
\usepackage{nicefrac}       
\usepackage{microtype}      
\usepackage[dvipsnames]{xcolor}         
\usepackage{amsmath}
\usepackage{amssymb}
\usepackage{mathtools}
\usepackage{makecell}
\usepackage{multirow}
\usepackage{enumitem}
\setlist[itemize]{
  noitemsep, topsep=0pt,
  label=$\blacktriangleright$,
  leftmargin=*
}
\usepackage[vlined, ruled, linesnumbered]{algorithm2e}
\usepackage{xspace}
\usepackage{pifont}
\newcommand{\xmark}{\textcolor{red}{\ding{55}}\xspace}%
\usepackage[color=Aquamarine!20, disable]{todonotes}
\usepackage{comment}
\usepackage{alltt}
\newcommand{\cond}[1]{{\bf \textsf{C#1}}\xspace}

\newcommand{\mclass}{\mathcal{M}}

\newcommand{\Loss}{\mathcal{L}}
\newcommand{\alg}{\mathcal{A}}
\newcommand{\falg}{\mathcal{F}}
\newcommand{\halg}{\mathcal{H}}
\newcommand{\btheta}{\boldsymbol{\theta}}
\newcommand{\bphi}{\boldsymbol{\phi}}
\newcommand{\bTheta}{\boldsymbol{\Theta}}
\newcommand{\bPhi}{\boldsymbol{\Phi}}

\newcommand{\R}{\mathbb{R}}

\title{FLoRA: Single-shot Hyper-parameter Optimization\\for Federated Learning}

\author{%
  Yi Zhou
   \qquad
   Parikshit Ram
   \qquad
   Theodoros Salonidis \\
   \And
   Nathalie Baracaldo 
   \qquad
   Horst Samulowitz
  \qquad
  Heiko Ludwig\\
  IBM Research\\
  \texttt{\{yi.zhou, Parikshit.Ram\}@ibm.com}\\
  \texttt{\{baracald, tsaloni, samulowitz, hludwig\}@us.ibm.com}
}

\begin{document}

\maketitle
\begin{abstract}
We address the relatively unexplored problem of hyper-parameter optimization (HPO) for federated learning (FL-HPO).  We introduce {\bf F}ederated {\bf Lo}ss Su{\bf R}face {\bf A}ggregation (FLoRA), the first FL-HPO solution framework that can address use cases of tabular data and gradient boosting training algorithms in addition to stochastic gradient descent/neural networks commonly addressed in the FL literature. The framework enables single-shot FL-HPO, by first identifying a good set of hyper-parameters that are used in a {\em single} FL training.  Thus, it enables FL-HPO solutions with minimal additional communication overhead compared to FL training without HPO.  Our  empirical evaluation of FLoRA for Gradient Boosted Decision Trees on seven OpenML data sets  demonstrates significant model accuracy improvements over the considered baseline, and robustness to increasing number of parties involved in FL-HPO training.

\end{abstract}
\section{Introduction} \label{sec:prob-setup}

Traditional machine learning (ML) approaches require training data to be gathered at a central location where the learning algorithm runs. In real world scenarios, however, training data is often subject to privacy or regulatory constraints restricting the way data can be shared, used and transmitted. Examples of such regulations include the European General Data Protection Regulation (GDPR), California Consumer Privacy Act (CCPA), Cybersecurity Law of China (CLA) and HIPAA, among others.
Federated learning (FL), first proposed in~\cite{mcmahan2017communication}, has recently become a popular approach to address privacy concerns by allowing collaborative training of ML models among multiple parties where each party can keep its data private.\\
\textbf{FL-HPO problem.} Despite the privacy protection FL brings along, there are many open problems in FL domain~\cite{kairouz2019advances,khodak2021federated}, one of which is hyper-parameter optimization for FL.
Existing FL systems require a user (or all participating parties) to pre-set (agree on) multiple hyper-parameters (HPs) (i) for the model being trained (such as number of layers and batch size for neural networks or tree depth and number of trees in tree ensembles), and (ii) for the the aggregator (if such hyper-parameters exist).  Hyper-parameter optimization (HPO) for FL is important because the choice of HPs can have dramatic impact on performance.  This is particularly important for tabular data (where datasets can be radically different from each other) as well as image data and neural nets.\\
While HPO has been widely studied in the centralized ML setting, it comes with unique challenges in the FL setting.  First, existing HPO techniques for centralized training often make use of the entire data set, which is not available in FL. Secondly, they train a vast variety of models for a large number of HP configurations which would be prohibitively expensive in terms of communication and training time in FL settings. Thirdly, one important challenge that has not been adequately explored in FL literature is support for tabular data, which are widely used in enterprise settings.
One of the best models for this setting are based on gradient boosting tree algorithms which are not based on the stochastic gradient descent training algorithm used for neural networks.  Recently, a few approaches have been proposed for FL-HPO,
however they focus on handling HPO using personalization techniques~\cite{khodak2021federated} and neural networks~\cite{khodak2020weight}. 
To the best of our knowledge, there is no HPO approach for FL systems to train non-neural network models, such as XGBoost that are particularly common in the enterprise setting.\\
\textbf{Scope.} 
In this paper, we address the aforementioned challenges of FL-HPO. 
We focus on the problem where the model HPs are shared across all parties
and we seek a set of HPs and train a single model that is eventually used by all parties for testing/deployment.
Moreover, we impose three further requirements that make the problem more challenging: 
(\cond{1}) 
we {\em do not make any assumption} that two models with different HPs can perform some form of ``weight-sharing'' (which is a common technique used in various HPO and neural architecture search (NAS) schemes for neural networks to reduce the computational overhead of HPO and NAS), allowing our solution to be applied beyond neural networks~\cite{khodak2020weight}. 
(\cond{2}) we seek to {\bf perform ``single-shot'' FL-HPO}, where we have {\em limited} resources (in the form of computation and communication overhead) which allow training only a single model via federated learning (that is, a single HP configuration), and 
(\cond{3}) we {\em do not assume that parties have independent and identically distributed (IID) data distributions}.\\ 
%
\textbf{Contributions.}
Given the above FL-HPO problem setting, we make the following contributions:
\begin{itemize}
\item (\S \ref{sec:method}) We present a novel framework {\em {\bf F}ederated {\bf Lo}ss Su{\bf R}face {\bf A}ggregation} (FLoRA) that leverages meta-learning techniques to utilize {\bf local and asynchronous} HPO on each party to perform single-shot HPO for the global FL-HPO problem. 
\item (\S \ref{sec:emp}) We evaluate FLoRA on the FL-HPO of Gradient Boosted Decision Trees (GBDTs)~\citep{friedman2001greedy} on seven classification datasets from OpenML~\citep{OpenML2013}, highlighting (i) its performance relative to the baseline, (ii) the effect of various choices in this scheme, (iii) the effect of the number of parties on the performance, and (iv) finally, we empirically demonstrate that FLoRA works under high heterogeneous (Non-IIDness) parties distributions. 
\end{itemize}


\section{Related work}\label{sec:related}
Despite fruitful research results on FL regarding advanced learning algorithms~\cite{karimireddy2019scaffold,ong2020adaptive}, data heterogeneity~\cite{zhao2018federated,li2018federated}, personalization~\cite{li2021ditto,smith2017federated,nichol2018first,fallah2020personalized}, fairness~\cite{abay2020mitigating,mohri2019agnostic}, system design~\cite{konevcny2016federated,chai2020tifl,bonawitz2019towards} and privacy-preserving frameworks\cite{agarwal2018cpsgd,truex2019hybrid,xu2019hybridalpha} etc., there are only a few focusing on hyper-parameters tuning for FL~\cite{Dai20FBO,khodak2020weight,Koskela19Learning,Mostafa19Robust,Reddi20Adaptive,khodak2021federated}. 
In~\cite{Dai20FBO}, Dai et.al. address Federated Bayesian Optimization. The problem setup is quite different than FL-HPO in that they focus on a single party using information from other parties to accelerate its own Bayesian Optimization. The works in~\cite{Koskela19Learning,Mostafa19Robust,Reddi20Adaptive} consider adaptation of learning rate in SGD-based FL training. 
Inspired from the neural architecture search technique of weight-sharing,~\cite{khodak2020weight, khodak2021federated} proposed FedEx, a FL-HPO framework to accelerate a general hyper-parameter tuning procedure, i.e., successive halving algorithm (SHA), for many SGD-based FL algorithms. 
In contrast to these approaches, our framework is also applicable to non SGD-training settings and minimizes HPO overhead by being a one-shot approach. 

While HPO has been widely studied in the centralized ML setting, it comes with unique challenges in the FL setting.  First, existing HPO techniques for centralized training often make use of the entire data set, which is not available in FL. Secondly, they train a vast variety of models for a large number of HP configurations which would be prohibitively expensive in terms of communication and tra
Beyond grid-search for HPO, random search is a very competitive baseline because of its simplicity and parallelizability \citep{bergstra2012random}. Sequential model-based optimization (SMBO) \citep{shahriari2016taking} is a common technique with different `surrogate models' such as Gaussian processes \citep{snoek2012practical}, random forests \citep{hutter2011sequential}, radial basis functions \citep{rbfopt}, and tree-parzen estimators \citep{bergstra2011algorithms}. However, black-box optimization is a time consuming process because the expensive black-box function evaluation involves model training and scoring (on a held-out set). Efficient {\em multi-fidelity} approximations of the black-box function based on some budget (training samples/epochs) combined with bandit learning can skip unpromising candidates early via successive halving \citep{jamieson2016non,sabharwal2016selecting} and  HyperBand \citep{li2018hyperband}. However, these schemes essentially perform an efficient random search and are well suited for search over discrete spaces or discretized continuous spaces. BOHB \citep{bohb} combines SMBO (with TPE) and HyperBand for improved optimization. The problem of HPO has been extended from ML model configurations to the configuration of complete ML pipelines --  the Combined Algorithm Selection and HPO (or CASH) problem -- with many increasingly efficient algorithms~\citep{autoweka1, feurer2015efficient, autoweka2, rakotoarison2019automated, liu2020admm}. All these techniques rely on multiple (partial or full) trainings of models with different HP configurations, and hence are not practical for the single-shot FL-HPO problem.\\
Instead of starting the HPO from scratch for every dataset (which is usually a very expensive process), meta-learning can be used to refine the search space and warm start the search  \citep{vanschoren2018meta}. The usual techniques involve ``meta-features'' for data sets that are used to develop a notion of similarity between data sets. This is then used for any new data set (on which HPO needs to be performed) to identify similar previously processed data sets~\citep{feurer2015efficient,wistuba2015learningDatasetSimilarities}. The HPO on previously processed data sets are also utilized to (i) generate promising initial HPs for the HPO on any new data set~\citep{feurer2015initializing, wistuba2015learningHPOinits}, and/or (ii) prune the HPO search space, removing unpromising areas, allowing the HPO solver to focus its attention on useful parts of the search space~\citep{wistuba2015hyperparameter,perrone2019learning}. For the purposes of FL-HPO, we can view the per-party data sets as ``similar'' data sets, even though we do not assume them to be IID, and explore how the meta-learning technique of learning good HPO initialization as a potential way of tackling single-shot FL-HPO.

\section{Methodology} \label{sec:method}

In the centralized ML setting, we would consider a model class $\mclass$ and its corresponding learning algorithm $\alg$ parameterized collectively with HPs $\btheta \in \bTheta$, and given a training set $D$, we can learn a single model $\alg(\mclass, \btheta, D) \to m \in \mclass$. Given some predictive loss $\Loss(m, D')$ of any model $m$ scored on some holdout set $D'$, the centralized HPO problem can be stated as
\begin{equation}
\label{eq:central-HPO}
\min\nolimits_{\btheta \in \bTheta} \Loss(\alg(\mclass, \btheta, D), D').
\end{equation}
In the most general FL setting, we have $p$ parties $P_1, \dots, P_p$ each with their private local training data set $D_i, i \in [p]$. Let $D = \cup_{i=1}^p D_i$ denote the aggregated training data set and $\overline{D} = \{D_i\}_{i \in [p]}$ denote the set of per-party data sets. Each model class (and corresponding learning algorithm) is parameterized by global HPs $\btheta_G \in \bTheta_G$ shared by all parties and per-party local HPs $\btheta_L^{(i)} \in \bTheta_L, i \in [p]$ with $\bTheta = \bTheta_G \times \bTheta_L$. FL systems usually include an aggregator with its own set of HPs $\bphi \in \bPhi$. Finally, we would have a FL algorithm 
$\falg \left( \mclass, \bphi, \btheta_G, \{ \btheta_L^{(i)} \}_{i \in [p]}, \alg, \overline{D} \right) \to m \in \mclass$ that takes as input all the relevant HPs and per-party data sets and generates a model. In this case, the FL-HPO problem can be stated in the two following ways depending on the desired goals: (i) For a global holdout data set $D'$ (a.k.a validation set, possibly from the same distribution as the aggregated data set $D$), we solve the following problem:
\begin{equation}
\label{eq:fl-hpo-1}
\min_{\bphi \in \bPhi, \btheta_G \in \bTheta_G, \btheta_L^{(i)} \in \bTheta_L, i \in [p]}
\Loss\left( \falg \left( 
\mclass, \bphi, \btheta_G, \{ \btheta_L^{(i)} \}_{i \in [p]}, \alg, \overline{D}
\right), D' \right).
\end{equation}
(ii) An alternative problem would involve per-party holdout data sets $D_i', i \in [p]$ and we solve the following problem:
\begin{equation}
\label{eq:fl-hpo-2}
\min_{\bphi \in \bPhi, \btheta_G \in \bTheta_G, \btheta_L^{(i)} \in \bTheta_L, i \in [p]}
\mathsf{Agg}\left(
\left\{
  \Loss\left( \falg \left(
  \mclass, \bphi, \btheta_G, \{ \btheta_L^{(i)} \}_{i \in [p]}, \alg, \overline{D}
  \right), D'_i \right),
  i \in [p] \right\}
\right),
\end{equation}
where $\mathsf{Agg}: \R^p \to \R$ is some aggregation function (such as average or maximum) that scalarizes the $p$ per-party predictive losses.

Contrasting problem \eqref{eq:central-HPO} to problems \eqref{eq:fl-hpo-1} \& \eqref{eq:fl-hpo-2}, we can see that the FL-HPO is significantly more complicated than the centralized HPO problem. In the ensuing presentation, we focus on problem~\eqref{eq:fl-hpo-1} although our proposed single-shot FL-HPO scheme can be applied and evaluated for problem~\eqref{eq:fl-hpo-2}. We simplify the FL-HPO problem in the following ways: (i) we assume that there is no personalization so there are no per-party local HPs $\btheta_L^{(i)}, i \in [p]$, and (ii) we only focus on the model class HPs $\btheta_G$, deferring HPO for aggregator HPs $\bphi$ for future work. Hence the problem we will study is stated as for a fixed aggregator HP $\bphi$:
\begin{equation}
\label{eq:fl-hpo-1a}
\min\nolimits_{\btheta_G \in \bTheta_G}
\Loss\left( \falg \left( 
\mclass, \bphi, \btheta_G, \alg, \overline{D}
\right), D' \right).
\end{equation}
This problem appears similar to the centralized HPO problem~\eqref{eq:central-HPO}. However, note that the main challenge in \eqref{eq:fl-hpo-1a} is the need for a federated training for each set of HPs $\btheta_G$, and hence it is not practical (from a communication overhead perspective) to apply existing off-the-shelf HPO schemes to problem~\eqref{eq:fl-hpo-1a}. In the subsequent discussion, for simplicity purposes, we will use $\btheta$ to denote the global HPs, dropping the ``$G$'' subscript.
%
\subsection{Leveraging local HPOs}
While it is impractical to apply off-the-shelf HPO solvers (such as Bayesian Optimization (BO)~\citep{shahriari2016taking}, Hyperopt~\citep{bergstra2011algorithms}, SMAC~\citep{hutter2011sequential}, and such), we wish to understand how we can leverage local and asynchronous HPOs in each of the parties. We begin with a simple but intuitive hypothesis underlying various meta-learning schemes for HPO~\citep{vanschoren2018meta,wistuba2018scalable}\todo{more citations}: {\em if a HP configuration $\btheta$ has good performance for all parties independently, then $\btheta$ is a strong candidate for federated training}.

%
\begin{algorithm}[h]
\DontPrintSemicolon
\SetAlgoLined
\caption{Single-shot FL-HPO with Federated Loss Surface Aggregation}
\label{alg:fl-hpo-lsa}
{\footnotesize
\SetKwProg{ssFLoRA}{FLoRA$(\bTheta, \mclass, \alg, \{ (D_i, D_i') \}_{i \in [p]}, T) \to m$}
{}
{end}
\ssFLoRA{}{
  \For{each party $P_i, i \in [p]$}{
    Run HPO to generate $T$ (HP, loss) pairs
    \begin{equation}\label{eq:local-hpo-data}
    E^{(i)} = \left \{
    (\btheta_t^{(i)}, \Loss_t^{(i)}), t \in [T],
    \btheta_t^{(i)} \in \bTheta,
    \Loss_t^{(i)} := \Loss(\alg(\mclass, \btheta_t^{(i)}, D_i), D_i') 
    \right \}
    \end{equation}
  }
  Collect all $E^{(i)}, i \in [p]$ in aggregator \;
  Generate a unified loss surface $\ell:\bTheta \to \R$ using $\left\{ E^{(i)}, i \in [p] \right\}$\;
  Select best HP candidate $\btheta^\star \gets \arg \min_{\btheta \in \bTheta} \ell(\btheta)$ \;
  Learn final model with federated training: $m \gets \falg(\mclass, \bphi, \btheta^\star, \alg, \overline{D})$\;
  \KwRet{$m$}
}
}
\end{algorithm}
With this hypothesis, we present our proposed algorithm {\bf FLoRA} in Algorithm~\ref{alg:fl-hpo-lsa}. In this scheme, we allow each party to perform HPO locally and asynchronously with some adaptive HPO scheme such as BO (line 3). Then, at each party $i \in [p]$, we collect all the attempted $T$ HPs $\btheta_t^{(i)}, t \in [T]$ and their corresponding predictive loss $\Loss_t^{(i)}$ into a set $E^{(i)}$ (line 3, equation~\eqref{eq:local-hpo-data}). Then these per-party sets of (HP, loss) pairs $E^{(i)}$ are collected at the aggregator (line 5). This operation has at most $O(pT)$ communication overhead (note that the number of HPs are usually much smaller than the number of columns or number of rows in the per-party data sets). These sets are then used to generate an aggregated loss surface $\ell: \bTheta \to \R$ (line 6) which will then be used to make the final single-shot HP recommendation $\btheta^\star \in \bTheta$ (line 7) for the federated training to create the final model $m \in \mclass$ (line 8). We will discuss the generation of the aggregated loss surface in detail in \S \ref{sec:method:lsa}. Before that, we briefly want to discuss the motivation behind some of our choices in Algorithm~\ref{alg:fl-hpo-lsa}.\\
\textbf{Why adaptive HPO?}
The reason to use adaptive HPO schemes instead of non-adaptive schemes such as random search or grid search is that this allows us to efficiently approximate the local loss surface more accurately (and with more certainty) in regions of the HP space where the local performance is favorable instead of trying to approximate the loss surface well over the complete HP space. This has advantages both in terms of computational efficiency and loss surface approximation.\\
\textbf{Why asynchronous HPO?}
Each party executes HPO asynchronously, without coordination with HPO results from other parties or with the aggregator. This is in line with our objective to minimize communication overhead. Although there could be strategies that involve coordination between parties, they could involve many rounds of communication. Our experimental results show that this approach is effective for the datasets we evaluated for.


\subsection{Loss surface aggregation} \label{sec:method:lsa}

Given the sets $E^{(i)}, i \in [p]$ of (HP, loss) pairs $(\btheta_t^{(i)}, \Loss_t^{(i)}), i \in [p], t \in [T]$ at the aggregator, we wish to construct a loss surface $\ell : \bTheta \to \R$ that best emulates the (relative) performance loss $\ell(\btheta)$ we would observe when training the model on $\overline{D}$. Based on our hypothesis, we want the loss surface to be such that it would have a relatively low $\ell(\btheta)$ if $\btheta$ has a low loss for all parties simultaneously. However, because of the asynchronous and adaptive nature of the local HPOs, for any HP $\btheta \in \bTheta$, we would not have the corresponding losses from all the parties. For that reason, we will model the loss surfaces using regressors that try to map any HP to their corresponding loss. In the following, we present four ways of constructing such loss surfaces:\\
\textbf{Single global model (SGM).}\todo{better name}
We merge all the sets $E^{(i)}, i \in [p]$ into $E$ and use it as a training set for a regressor $f: \bTheta \to \R$, which considers the HPs $\btheta \in \bTheta$ as the covariates and the corresponding loss as the dependent variable. For example, we can train a random forest regressor\todo{add citation} on this training set $E$. Then we can define the loss surface $\ell(\btheta) := f(\btheta)$. However, this loss surface does not have our desirable properties: it is actually overly optimistic -- under the assumption that every party generates unique HPs during the local HPO, this single global loss surface would assign a low loss to any HP $\btheta$ which has a low loss at any one of the parties. This implies that this loss surface would end up recommending HPs that have low loss in just one of the parties.\\
\textbf{Single global model with uncertainty (SGM+U).}
Given the merged set $E$ of the per-party sets of (HP, loss) pairs, we can train a regressor that provides uncertainty quantification around its predictions (such as Gaussian Process Regressor\todo{add citation}) as $f: \bTheta \to \R, u: \bTheta \to \R_+$, where $f(\btheta)$ is the mean prediction of the model at $\btheta \in \bTheta$ while $u(\btheta)$ quantifies the uncertainty around this prediction $f(\btheta)$. We define the loss surface as $\ell(\btheta) := f(\btheta) + \alpha \cdot u(\btheta)$ for some $\alpha > 0$. This loss surface does prefer HPs that have a low loss even in just one of the parties, but it penalizes a HP if the model estimates high uncertainty around this HP. Usually, a high uncertainty around a HP would be either because the training set $E$ does not have many samples around this HP (implying that not many parties thought that the region where this HP lies is a region for low loss), or because there are multiple samples in the region around this HP but parties do not collectively agree that this is a promising region for HPs. Hence this makes SGM+U more desirable than SGM, giving us a loss surface that estimates low loss for HPs that are simultaneously thought to be promising to multiple parties.\\
\textbf{Maximum of per-party local models (MPLM).}
Instead of a single global model on the merged set $E$, we can instead train a regressor $f^{(i)}: \bTheta \to \R, i \in [p]$ with each of the per-party set $E^{(i)}, i \in [p]$ of (HP, loss) pairs. Given this, we can construct the loss surface as $\ell(\btheta) := \max_{i \in [p]} f^{(i)}(\btheta)$. This can be seen as a much more pessimistic loss surface, assigning a low loss to a HP only if it has a low loss estimate across all parties. \\
\textbf{Average of per-party local models (APLM).}
A less pessimistic version of MPLM would be to construct the loss surface as the average of the per-party regressors $f^{(i)}, i \in [p]$ instead of the maximum, defined as $\ell(\btheta) := \nicefrac{1}{p} \sum_{i=1}^p f^{(i)}(\btheta)$. This is also less optimistic than SGM since it will assign a low loss for a HP only if its average across all per-party regressors is low, which implies that all parties observed a relatively low loss around this HP.\\
Intuitively, we believe that loss surfaces such as SGM+U or APLM would be the most promising while the extremely optimistic and pessimistic  SGM and MPLM respectively would be relatively less promising, with MPLM being superior to SGM. In the following section, we evaluate all these loss surface empirically in the single-shot FL-HPO setting.

\section{Empirical evaluation} \label{sec:emp}

\begin{table}[t]
\centering
\caption{Comparison of different loss surfaces (the 4 rightmost columns)  for FLoRA relative to the  baseline for single-shot 3-party FL-HPO in terms of the {\em relative regret} (lower is better). See text in \S \ref{sec:emp:ssb} for the detailed description of ``Party max/min''.
}
{\scriptsize
\begin{tabular}{lccccc}
\toprule
Data          & Party max/min & SGM & SGM+U & MPLM & APLM \\
\midrule
EEG eye state & 1.005 & 0.1507 & 0.1347 & {\bf 0.1233} & 0.1279 \\
Electricity   & 1.009 & 0.1848 & 0.1518 & {\bf 0.1089} & 0.1381 \\
Heart statlog & 1.109 & 0.6904 & 0.5543 & 0.8930 & {\bf 0.5008} \\
Oil spill     & 1.205 & 0.7086 & {\bf 0.4032} & 0.5678 & 0.5282 \\
PC3           & 1.044 & 0.6639 & 0.7220 & 0.3921 & {\bf 0.3797} \\
Pollen        & 1.016 & 0.4328 & 0.5403 & {\bf 0.4269} & 0.6896 \\
Sonar         & 1.055 & 1.3298 & {\bf 0.4058} & 0.9215 & 0.7094 \\
\midrule
Aggregate     &  -    & 0.5944 $\pm$ 0.3997 & {\bf 0.416 $\pm$ 0.21} & 0.4905 $\pm$ 0.3286 & 0.4391 $\pm$ 0.2375 \\
\bottomrule
\end{tabular}
}
\label{tab:baseline-comp}
\end{table}

In this section, we evaluate our proposed scheme and different loss surfaces for the FL-HPO of gradient boosted decision trees~\citep{friedman2001greedy} on OpenML~\citep{OpenML2013} classification problems. Specifically, we focus on the histogram based gradient boosting, available as {\tt HistGradientBoostingClassifier} in the {\tt sklearn.ensemble} module of the {\tt scikit-learn} library~\citep{scikit-learn}. The precise HP search space is described in Appendix~\ref{asec:emp:search-space}. First, we fix the number of parties $p = 3$ and compare our proposed scheme to a baseline on $7$ data sets. Then we study the effect of increasing the number of parties from $p = 3$ to $p = 10$ on the performance of our proposed scheme on $3$ data sets. The data is randomly split across parties.
Finally, we evaluate our proposed {\bf FLoRA} in a real FL testbed IBM FL~\cite{ludwig2020ibm} using its default HP setting as a baseline.\\
\textbf{Single-shot baseline.}
To appropriately evaluate our proposed single-shot FL-HPO scheme, we need to select a meaningful single-shot baseline. For this, we choose the default HP configuration of {\tt HistGradientBoostingClassifier} in {\tt scikit-learn} as the single-shot baseline. We choose this baseline for two main reasons: (i) this default HP configuration in {\tt scikit-learn} is set manually based on expert prior knowledge and extensive empirical evaluation\todo{add citation}, and (ii) this HP configuration is also used as the default for gradient boosting decision trees in the Auto-Sklearn package~\citep{feurer2015efficient,feurer-arxiv20a}, one of the leading open-source AutoML python packages, which maintains a carefully selected portfolio of default configurations.\\
\textbf{Data set selection.}
For our evaluation of single-shot HPO, we consider 7 binary classification data sets of varying sizes and characteristics from OpenML~\citep{OpenML2013} such that there is at least a significant room for improvement over the performance single-shot baseline. We consider data sets which have at least $> 3\%$ potential improvement in balanced accuracy. See Appendix~\ref{asec:emp:datasets} for details on data.\\
\textbf{Implementation.}
We consider two implementations for our empirical evaluation. In our first set of experiments in \S \ref{sec:emp:ssb} and \S \ref{sec:emp:np}, we emulate the final FL (Algorithm~\ref{alg:fl-hpo-lsa}, line 8) with a centralized training using the pooled data. We chose this implementation because we want to evaluate the final performance of any HP configuration (baseline or recommended by FLoRA) in a statistically robust manner with multiple train/validation splits (for example, via 10-fold cross-validation) instead of evaluating the performance on a single train/validation. This form of evaluation is extremely expensive to perform in a real FL system. This form of evaluation also allows us to evaluate how the performance of our single-shot HP recommendation fairs against that of the best-possible HP found via a full-scale centralized HPO. This is again not feasible in a real FL system. To highlight that our proposed scheme do translate to improved performance in a real FL testbed, we utilize the IBM FL library~\citep{ludwig2020ibm} on 3 of the data sets in \S \ref{sec:emp:flexpts}. In that case, we report the metrics on a single train/test split.\\
\textbf{Evaluation metric.}
In all data sets, we consider the balanced accuracy as the metric we wish to maximize. For the local per-party HPOs (as well as the centralized HPO we execute to compute the regret), we maximize the 10-fold cross-validated balanced accuracy. For the experiments in \S \ref{sec:emp:ssb} and \S \ref{sec:emp:np}, we report the relative regret, computed as $\nicefrac{(a^\star - a)}{(a^\star - b)}$, where $a^\star$ is the best possible accuracy obtained via the centralized HPO, $b$ is the accuracy of the baseline, and $a$ is the accuracy of the HP recommended by any scheme (baseline or FLoRA). The baseline has a relative regret of 1 and smaller values imply better performance. A value larger than 1 implies that the recommended HP performs worse than the baseline. For the experiments in \S \ref{sec:emp:flexpts} with a real FL system, we report the balanced accuracy of any HP (baseline or recommended by FLoRA) on a single train/test split. Given balanced accuracy as the evaluation metric, we utilize (1 - balanced accuracy) as the loss $\Loss_t^{(i)}$ in Algorithm~\ref{alg:fl-hpo-lsa}.

\subsection{Comparison to single-shot baseline} \label{sec:emp:ssb}
In our first set of experiments for 3-party FL-HPO ($p = 3$), we compare our proposed scheme with the baseline across different data sets and report the relative regret for different choices of the loss surfaces in Table~\ref{tab:baseline-comp}. In this table, we also report the following ratio in the second column as the ``Party max/min''\todo{better name}: $\nicefrac{\left(1 - \min_{i \in [p]} \Loss_\star^{(i)} \right) }{ \left(1 - \max_{i \in [p]} \Loss_\star^{(i)} \right)}$, where $\Loss_\star^{(i)} = \min_{t \in [T]} \Loss_t^{(i)}$ is the minimum loss observed during the local asynchronous HPO at party $i$. This ratio is always greater than 1, and highlights the difference in the observed performances across the parties. A ratio closer to 1 indicates that all the parties have relatively similar performances on their training data, while a ratio much higher than 1 indicating significant discrepancy between the per-party performances, implicitly indicating the difference in the per-party data distributions. Table~\ref{tab:baseline-comp} indicates that there are significant differences for Oil spill and Heart statlog data sets and very small differences for the Electricity and EEG eye state data sets.\\
The results indicate that, in almost all cases, with all loss functions, our proposed scheme is able to improve upon the baseline to varying degrees (there is only one case where SGM performs worse than the baseline on Sonar). On average (across the data sets), SGM+U and APLM perform the best as we expected, with both of them also having significantly smaller standard deviations for the relative regret compared to SGM and MPLM. MPLM performs better than SGM both in terms of average and standard deviation. Looking at the individual data sets, we see that, for data sets with low ``Party max/min'' (EEG eye state, Electricity), all the proposed loss surface have low relative regret, indicating that the problem is easier as expected. For data sets with high ``Party max/min'' (Heart statlog, Oil spill), the relative regret of all loss surfaces are higher (but still much smaller than 1), indicating that our proposed single-shot scheme can show improvement even in cases where there is significant difference in the per-party losses (and hence data sets).


\begin{table}[t]
\centering
\caption{Effect of increasing the number of parties on FLoRA with different loss surfaces. The experimental setup is described in \S \ref{sec:emp:np}.}
{\scriptsize
\begin{tabular}{lcccccc}
\toprule
Data          & \# parties & Party max/min & SGM & SGM+U & MPLM & APLM \\
\midrule
EEG eye state &  3       & 1.005 & 0.1507 & 0.1347 & {\bf 0.1233} & 0.1279 \\
14980 rows    &  6       & 1.011 & 0.0685 & {\bf 0.0023} & 0.0753 & 0.0890 \\
              & 10       & 1.033 & 0.0822 & {\bf 0.0000} & 0.1644 & 0.0137 \\
\midrule
Electricity   &  3       & 1.009 & 0.1848 & 0.1518 & {\bf 0.1089} & 0.1381 \\
45312 rows    &  6       & 1.007 & 0.2626 & 0.2198 & 0.1907 & {\bf 0.1420} \\
              & 10       & 1.005 & {\bf 0.0447} & 0.0700 & 0.3385 & 0.1518 \\
\midrule
Pollen        &  3       & 1.016 &  0.4328 & 0.5403 & {\bf 0.4269} & 0.6896 \\
3848 rows     &  6       & 1.101 & 1.0239 & 0.9164 & {\bf 0.5403} & 0.5644 \\
              & 10       & 1.159 & 1.0478 & {\bf 0.7313} & 0.7522 & 1.1254 \\
\bottomrule
\end{tabular}
}
\label{tab:num-parties-comp}
\end{table}

\subsection{Effect of increasing number of parties} \label{sec:emp:np}
In the second set of experiments, we study the effect of increasing the number of parties in the FL-HPO problem on 3 data sets. We present the relative regrets (along with the ``Party max/min'') in Table~\ref{tab:num-parties-comp}. We notice that increasing the number of parties does not have a significant effect on the ``Party max/min'' for the Electricity data set, but significantly increases for the Pollen data set (making the problem harder). For the EEG eye state, the increase in the ``Party max/min'' with increasing number of parties is moderate. 
\linebreak
The results indicate that, with low or moderate increase in ``Party max/min'' (EEG eye state, Electricity), the proposed scheme is able to achieve low relative regret -- the increase in the number of parties does not directly imply degradation in performance. However, with significant increase in ``Party max/min'' (Pollen), we see a significant increase in the relative regret (eventually going over 1 in a few cases). The difference in data size with Pollen having far fewer data points than the others is likely to have an impact on the performance as well. It is important to note that in this challenging case, MPLM (the most pessimistic loss function) has the most graceful degradation in relative regret compared to the remaining loss surfaces.
%
\subsection{Federated Learning testbed evaluation} \label{sec:emp:flexpts}
We now conduct experiments in a FL testbed, utilizing IBM FL library~\citep{ludwig2020ibm},
which contains an implementation of {\tt HistGradientBoostingClassifier} for FL~\cite{ong2020adaptive}. More specifically, we reserved $40\%$ of oil spill and electricity and $20\%$ of EEG eye state as hold-out test set to evaluate the final FL model performance while each party randomly sampled from the rest of the original dataset to obtain their own training dataset. 
We use the same HP search space as in Appendix~\ref{asec:emp:search-space}. Our target metric for all experiments is balanced accuracy. 
Each party will run HPO to generate $T = 500$ (HP, loss) pairs and use those pairs to generate loss surface either collaboratively or by their own according to different aggregation procedures described in \S \ref{sec:method:lsa}.
Once the loss surface is generated, the aggregator uses Hyperopt~\cite{bergstra2011algorithms} to select the best HP candidate and train a federated XGBoost model via the IBM FL library using the selected HPs.
Table~\ref{tab:fl-summary}
summarizes the experimental results for $3$ datasets, indicating that FLoRA can significantly improve over the baseline in IBM FL testbed. 

\begin{table}[h]
\centering
\caption{Performance of FLoRA with the IBM-FL system in terms of the {\em balanced accuracy} on a holdout test set (higher is better). The baseline is still the default HP configuration of {\tt HistGradientBoostingClassifier} in {\tt scikit-learn}.}
{\scriptsize
\begin{tabular}{lccccccc}
\toprule
Data          & \# parties & \# training data per party & Baseline & SGM & SGM+U & MPLM & APLM  \\
\midrule
Oil spill       &  $3$   &$200$   & 0.5895 & \textbf{0.7374} & 0.5909 & 0.7061 & 0.7332 \\
\midrule
EEG eye state &  $3$ & $3,000$  & 0.8864 & 0.9153 & 0.9211 & \textbf{0.9251} & 0.9245 \\
              
\midrule
Electricity   &  $6$  & $4,000$     & 0.8448 & 0.8562 & \textbf{0.8627} & 0.8621 & 0.8624 \\
             
\bottomrule
\end{tabular}
}
\label{tab:fl-summary}
\end{table}

\section{Conclusions and next steps} \label{sec:next-steps}
How to effectively select HP in FL settings is an unsolved problem.
In this paper, we introduced FLoRA, a single-shot FL-HPO algorithm that can be applied to a variety of ML models.
Our experimental evaluation shows that FLoRA can produce HPO configurations that outperform the baseline and deal with highly heterogeneous distributions among parties.
We plan to evaluate FLoRA on more data sets and FL-HPO of more machine learning methods (such as random decision forests, nearest neighbor models, kernel machines, neural network) to further quantify its performance. Moreover, we plan to extend our proposed algorithm to go from single-shot (one HP recommendation for one FL training) to a few-shot setup (where we would be allowed to perform a very small number of FL trainings). Finally, we plan to extend this approach to allow for personalization, using local party-specific HPs.
{
\small
\bibliographystyle{unsrtnat}
\bibliography{reference}

\begin{thebibliography}{51}
\providecommand{\natexlab}[1]{#1}
\providecommand{\url}[1]{\texttt{#1}}
\expandafter\ifx\csname urlstyle\endcsname\relax
  \providecommand{\doi}[1]{doi: #1}\else
  \providecommand{\doi}{doi: \begingroup \urlstyle{rm}\Url}\fi

\bibitem[McMahan et~al.(2017)McMahan, Moore, Ramage, Hampson, and
  y~Arcas]{mcmahan2017communication}
Brendan McMahan, Eider Moore, Daniel Ramage, Seth Hampson, and Blaise~Aguera
  y~Arcas.
\newblock Communication-efficient learning of deep networks from decentralized
  data.
\newblock In \emph{Artificial intelligence and statistics}, pages 1273--1282.
  PMLR, 2017.

\bibitem[Kairouz et~al.(2019)Kairouz, McMahan, Avent, Bellet, Bennis, Bhagoji,
  Bonawitz, Charles, Cormode, Cummings, et~al.]{kairouz2019advances}
Peter Kairouz, H~Brendan McMahan, Brendan Avent, Aur{\'e}lien Bellet, Mehdi
  Bennis, Arjun~Nitin Bhagoji, Kallista Bonawitz, Zachary Charles, Graham
  Cormode, Rachel Cummings, et~al.
\newblock Advances and open problems in federated learning.
\newblock \emph{arXiv preprint arXiv:1912.04977}, 2019.

\bibitem[Khodak et~al.(2021)Khodak, Tu, Li, Li, Balcan, Smith, and
  Talwalkar]{khodak2021federated}
Mikhail Khodak, Renbo Tu, Tian Li, Liam Li, Maria-Florina Balcan, Virginia
  Smith, and Ameet Talwalkar.
\newblock Federated hyperparameter tuning: Challenges, baselines, and
  connections to weight-sharing.
\newblock \emph{arXiv preprint arXiv:2106.04502}, 2021.

\bibitem[Khodak et~al.(2020)Khodak, Li, Li, Balcan, Smith, and
  Talwalkar]{khodak2020weight}
Mikhail Khodak, Tian Li, Liam Li, M~Balcan, Virginia Smith, and Ameet
  Talwalkar.
\newblock Weight sharing for hyperparameter optimization in federated learning.
\newblock In \emph{Int. Workshop on Federated Learning for User Privacy and
  Data Confidentiality in Conjunction with ICML 2020}, 2020.

\bibitem[Friedman(2001)]{friedman2001greedy}
Jerome~H Friedman.
\newblock Greedy function approximation: a gradient boosting machine.
\newblock \emph{Annals of statistics}, pages 1189--1232, 2001.

\bibitem[Vanschoren et~al.(2013)Vanschoren, van Rijn, Bischl, and
  Torgo]{OpenML2013}
Joaquin Vanschoren, Jan~N. van Rijn, Bernd Bischl, and Luis Torgo.
\newblock {OpenML}: Networked science in machine learning.
\newblock \emph{SIGKDD Explorations}, 15\penalty0 (2):\penalty0 49--60, 2013.
\newblock \doi{10.1145/2641190.2641198}.
\newblock URL \url{http://doi.acm.org/10.1145/2641190.2641198}.

\bibitem[Karimireddy et~al.(2019)Karimireddy, Kale, Mohri, Reddi, Stich, and
  Suresh]{karimireddy2019scaffold}
Sai~Praneeth Karimireddy, Satyen Kale, Mehryar Mohri, Sashank~J Reddi,
  Sebastian~U Stich, and Ananda~Theertha Suresh.
\newblock Scaffold: Stochastic controlled averaging for on-device federated
  learning.
\newblock 2019.

\bibitem[Ong et~al.(2020)Ong, Zhou, Baracaldo, and Ludwig]{ong2020adaptive}
Yuya~Jeremy Ong, Yi~Zhou, Nathalie Baracaldo, and Heiko Ludwig.
\newblock Adaptive histogram-based gradient boosted trees for federated
  learning.
\newblock \emph{arXiv preprint arXiv:2012.06670}, 2020.

\bibitem[Zhao et~al.(2018)Zhao, Li, Lai, Suda, Civin, and
  Chandra]{zhao2018federated}
Yue Zhao, Meng Li, Liangzhen Lai, Naveen Suda, Damon Civin, and Vikas Chandra.
\newblock Federated learning with non-iid data.
\newblock \emph{arXiv preprint arXiv:1806.00582}, 2018.

\bibitem[Li et~al.(2018{\natexlab{a}})Li, Sahu, Zaheer, Sanjabi, Talwalkar, and
  Smith]{li2018federated}
Tian Li, Anit~Kumar Sahu, Manzil Zaheer, Maziar Sanjabi, Ameet Talwalkar, and
  Virginia Smith.
\newblock Federated optimization in heterogeneous networks.
\newblock \emph{arXiv preprint arXiv:1812.06127}, 2018{\natexlab{a}}.

\bibitem[Li et~al.(2021)Li, Hu, Beirami, and Smith]{li2021ditto}
Tian Li, Shengyuan Hu, Ahmad Beirami, and Virginia Smith.
\newblock Ditto: Fair and robust federated learning through personalization.
\newblock In \emph{International Conference on Machine Learning}, pages
  6357--6368. PMLR, 2021.

\bibitem[Smith et~al.(2017)Smith, Chiang, Sanjabi, and
  Talwalkar]{smith2017federated}
Virginia Smith, Chao-Kai Chiang, Maziar Sanjabi, and Ameet Talwalkar.
\newblock Federated multi-task learning.
\newblock \emph{arXiv preprint arXiv:1705.10467}, 2017.

\bibitem[Nichol et~al.(2018)Nichol, Achiam, and Schulman]{nichol2018first}
Alex Nichol, Joshua Achiam, and John Schulman.
\newblock On first-order meta-learning algorithms.
\newblock \emph{arXiv preprint arXiv:1803.02999}, 2018.

\bibitem[Fallah et~al.(2020)Fallah, Mokhtari, and
  Ozdaglar]{fallah2020personalized}
Alireza Fallah, Aryan Mokhtari, and Asuman Ozdaglar.
\newblock Personalized federated learning with theoretical guarantees: A
  model-agnostic meta-learning approach.
\newblock \emph{Advances in Neural Information Processing Systems},
  33:\penalty0 3557--3568, 2020.

\bibitem[Abay et~al.(2020)Abay, Zhou, Baracaldo, Rajamoni, Chuba, and
  Ludwig]{abay2020mitigating}
Annie Abay, Yi~Zhou, Nathalie Baracaldo, Shashank Rajamoni, Ebube Chuba, and
  Heiko Ludwig.
\newblock Mitigating bias in federated learning.
\newblock \emph{arXiv preprint arXiv:2012.02447}, 2020.

\bibitem[Mohri et~al.(2019)Mohri, Sivek, and Suresh]{mohri2019agnostic}
Mehryar Mohri, Gary Sivek, and Ananda~Theertha Suresh.
\newblock Agnostic federated learning.
\newblock In \emph{International Conference on Machine Learning}, pages
  4615--4625. PMLR, 2019.

\bibitem[Kone{\v{c}}n{\`y} et~al.(2016)Kone{\v{c}}n{\`y}, McMahan, Yu,
  Richt{\'a}rik, Suresh, and Bacon]{konevcny2016federated}
Jakub Kone{\v{c}}n{\`y}, H~Brendan McMahan, Felix~X Yu, Peter Richt{\'a}rik,
  Ananda~Theertha Suresh, and Dave Bacon.
\newblock Federated learning: Strategies for improving communication
  efficiency.
\newblock \emph{arXiv preprint arXiv:1610.05492}, 2016.

\bibitem[Chai et~al.(2020)Chai, Ali, Zawad, Truex, Anwar, Baracaldo, Zhou,
  Ludwig, Yan, and Cheng]{chai2020tifl}
Zheng Chai, Ahsan Ali, Syed Zawad, Stacey Truex, Ali Anwar, Nathalie Baracaldo,
  Yi~Zhou, Heiko Ludwig, Feng Yan, and Yue Cheng.
\newblock Tifl: A tier-based federated learning system.
\newblock In \emph{Proceedings of the 29th International Symposium on
  High-Performance Parallel and Distributed Computing}, pages 125--136, 2020.

\bibitem[Bonawitz et~al.(2019)Bonawitz, Eichner, Grieskamp, Huba, Ingerman,
  Ivanov, Kiddon, Kone{\v{c}}n{\`y}, Mazzocchi, McMahan,
  et~al.]{bonawitz2019towards}
Keith Bonawitz, Hubert Eichner, Wolfgang Grieskamp, Dzmitry Huba, Alex
  Ingerman, Vladimir Ivanov, Chloe Kiddon, Jakub Kone{\v{c}}n{\`y}, Stefano
  Mazzocchi, H~Brendan McMahan, et~al.
\newblock Towards federated learning at scale: System design.
\newblock \emph{arXiv preprint arXiv:1902.01046}, 2019.

\bibitem[Agarwal et~al.(2018)Agarwal, Suresh, Yu, Kumar, and
  Mcmahan]{agarwal2018cpsgd}
Naman Agarwal, Ananda~Theertha Suresh, Felix Yu, Sanjiv Kumar, and H~Brendan
  Mcmahan.
\newblock cpsgd: Communication-efficient and differentially-private distributed
  sgd.
\newblock \emph{arXiv preprint arXiv:1805.10559}, 2018.

\bibitem[Truex et~al.(2019)Truex, Baracaldo, Anwar, Steinke, Ludwig, Zhang, and
  Zhou]{truex2019hybrid}
Stacey Truex, Nathalie Baracaldo, Ali Anwar, Thomas Steinke, Heiko Ludwig, Rui
  Zhang, and Yi~Zhou.
\newblock A hybrid approach to privacy-preserving federated learning.
\newblock In \emph{Proceedings of the 12th ACM Workshop on Artificial
  Intelligence and Security}, pages 1--11, 2019.

\bibitem[Xu et~al.(2019)Xu, Baracaldo, Zhou, Anwar, and
  Ludwig]{xu2019hybridalpha}
Runhua Xu, Nathalie Baracaldo, Yi~Zhou, Ali Anwar, and Heiko Ludwig.
\newblock Hybridalpha: An efficient approach for privacy-preserving federated
  learning.
\newblock In \emph{Proceedings of the 12th ACM Workshop on Artificial
  Intelligence and Security}, pages 13--23, 2019.

\bibitem[Dai et~al.(2020)Dai, Low, and Jaillet]{Dai20FBO}
Z.~Dai, B.K.H. Low, and P.~Jaillet.
\newblock Federated bayesian optimization via thompson sampling.
\newblock \emph{Advances in Neural Information Processing Systems}, 33, 2020.

\bibitem[Koskela and Honkela(2019)]{Koskela19Learning}
A.~Koskela and A.~Honkela.
\newblock Learning rate adaptation for federated and differentially private
  learning.
\newblock \emph{arXiv preprint arXiv:1809.03832}, 2019.

\bibitem[Mostafa(2019)]{Mostafa19Robust}
H.~Mostafa.
\newblock Robust federated learning through representation matching and
  adaptive hyper-parameters.
\newblock \emph{arXiv preprint arXiv:1912.13075}, 2019.

\bibitem[Reddi et~al.(2020)Reddi, Charles, Zaheer, Garrett, Rush, Konecny,
  Kumar, and McMahan]{Reddi20Adaptive}
S.J. Reddi, Z.~Charles, M.~Zaheer, Z.~Garrett, K.~Rush, J.~Konecny, S.~Kumar,
  and H.B. McMahan.
\newblock Adaptive federated optimization.
\newblock In \emph{International Conference on Learning Representations}, 2020.

\bibitem[Bergstra and Bengio(2012)]{bergstra2012random}
James Bergstra and Yoshua Bengio.
\newblock Random search for hyper-parameter optimization.
\newblock \emph{Journal of Machine Learning Research}, 13\penalty0
  (Feb):\penalty0 281--305, 2012.

\bibitem[Shahriari et~al.(2016)Shahriari, Swersky, Wang, Adams, and
  De~Freitas]{shahriari2016taking}
B.~Shahriari, K.~Swersky, Z.~Wang, R.~P. Adams, and N.~De~Freitas.
\newblock Taking the human out of the loop: A review of bayesian optimization.
\newblock \emph{Proceedings of the IEEE}, 104\penalty0 (1):\penalty0 148--175,
  2016.

\bibitem[Snoek et~al.(2012)Snoek, Larochelle, and Adams]{snoek2012practical}
J.~Snoek, H.~Larochelle, and R.~P. Adams.
\newblock Practical bayesian optimization of machine learning algorithms.
\newblock In \emph{Advances in neural information processing systems}, 2012.

\bibitem[Hutter et~al.(2011)Hutter, Hoos, and
  Leyton-Brown]{hutter2011sequential}
Frank Hutter, Holger~H Hoos, and Kevin Leyton-Brown.
\newblock Sequential model-based optimization for general algorithm
  configuration.
\newblock In \emph{International Conference on Learning and Intelligent
  Optimization}, pages 507--523. Springer, 2011.

\bibitem[Costa and Nannicini(2018)]{rbfopt}
A~Costa and G~Nannicini.
\newblock Rbfopt: an open-source library for black-box optimization with costly
  function evaluations.
\newblock \emph{Math. Prog. Comp. 10}, 2018.

\bibitem[Bergstra et~al.(2011)Bergstra, Bardenet, Bengio, and
  K{\'e}gl]{bergstra2011algorithms}
James~S Bergstra, R{\'e}mi Bardenet, Yoshua Bengio, and Bal{\'a}zs K{\'e}gl.
\newblock Algorithms for hyper-parameter optimization.
\newblock In \emph{Advances in neural information processing systems}, pages
  2546--2554, 2011.

\bibitem[Jamieson and Talwalkar(2016)]{jamieson2016non}
Kevin Jamieson and Ameet Talwalkar.
\newblock Non-stochastic best arm identification and hyperparameter
  optimization.
\newblock In \emph{Artificial Intelligence and Statistics}, pages 240--248,
  2016.

\bibitem[Sabharwal et~al.(2016)Sabharwal, Samulowitz, and
  Tesauro]{sabharwal2016selecting}
Ashish Sabharwal, Horst Samulowitz, and Gerald Tesauro.
\newblock Selecting near-optimal learners via incremental data allocation.
\newblock In \emph{Thirtieth AAAI Conference on Artificial Intelligence}, 2016.

\bibitem[Li et~al.(2018{\natexlab{b}})Li, Jamieson, DeSalvo, Rostamizadeh, and
  Talwalkar]{li2018hyperband}
Lisha Li, Kevin Jamieson, Giulia DeSalvo, Afshin Rostamizadeh, and Ameet
  Talwalkar.
\newblock Hyperband: A novel bandit-based approach to hyperparameter
  optimization.
\newblock \emph{Journal of Machine Learning Research}, 18\penalty0
  (185):\penalty0 1--52, 2018{\natexlab{b}}.

\bibitem[Falkner et~al.(2018)Falkner, Klein, and Hutter]{bohb}
Stefan Falkner, Aaron Klein, and Frank Hutter.
\newblock {BOHB}: Robust and efficient hyperparameter optimization at scale.
\newblock In \emph{Proceedings of the 35th International Conference on Machine
  Learning}, pages 1437--1446, 2018.

\bibitem[Thornton et~al.(2012)Thornton, Hoos, Hutter, and
  Leyton-Brown]{autoweka1}
Chris Thornton, Holger~H. Hoos, Frank Hutter, and Kevin Leyton-Brown.
\newblock Auto-weka: Automated selection and hyper-parameter optimization of
  classification algorithms.
\newblock \emph{arXiv}, 2012.
\newblock URL \url{http://arxiv.org/abs/1208.3719}.

\bibitem[Feurer et~al.(2015{\natexlab{a}})Feurer, Klein, Eggensperger,
  Springenberg, Blum, and Hutter]{feurer2015efficient}
Matthias Feurer, Aaron Klein, Katharina Eggensperger, Jost Springenberg, Manuel
  Blum, and Frank Hutter.
\newblock Efficient and robust automated machine learning.
\newblock In \emph{Advances in Neural Information Processing Systems}, pages
  2962--2970, 2015{\natexlab{a}}.

\bibitem[Kotthoff et~al.(2017)Kotthoff, Thornton, Hoos, Hutter, and
  Leyton-Brown]{autoweka2}
Lars Kotthoff, Chris Thornton, Holger~H. Hoos, Frank Hutter, and Kevin
  Leyton-Brown.
\newblock Auto-weka 2.0: Automatic model selection and hyperparameter
  optimization in weka.
\newblock \emph{J. Mach. Learn. Res.}, 18\penalty0 (1):\penalty0 826--830,
  January 2017.
\newblock ISSN 1532-4435.
\newblock URL \url{http://dl.acm.org/citation.cfm?id=3122009.3122034}.

\bibitem[Rakotoarison et~al.(2019)Rakotoarison, Schoenauer, and
  Sebag]{rakotoarison2019automated}
Herilalaina Rakotoarison, Marc Schoenauer, and Michele Sebag.
\newblock Automated machine learning with monte-carlo tree search.
\newblock In \emph{Proceedings of the Twenty-Eighth International Joint
  Conference on Artificial Intelligence, {IJCAI-19}}, pages 3296--3303, 2019.

\bibitem[Liu et~al.(2020)Liu, Ram, Vijaykeerthy, Bouneffouf, Bramble,
  Samulowitz, Wang, Conn, and Gray]{liu2020admm}
Sijia Liu, Parikshit Ram, Deepak Vijaykeerthy, Djallel Bouneffouf, Gregory
  Bramble, Horst Samulowitz, Dakuo Wang, Andrew Conn, and Alexander Gray.
\newblock An {ADMM} based framework for automl pipeline configuration.
\newblock In \emph{Thirty-Fourth AAAI Conference on Artificial Intelligence},
  2020.
\newblock URL \url{https://arxiv.org/abs/1905.00424v5}.

\bibitem[Vanschoren(2018)]{vanschoren2018meta}
Joaquin Vanschoren.
\newblock Meta-learning: A survey.
\newblock \emph{arXiv preprint arXiv:1810.03548}, 2018.

\bibitem[Wistuba et~al.(2015{\natexlab{a}})Wistuba, Schilling, and
  Schmidt-Thieme]{wistuba2015learningDatasetSimilarities}
Martin Wistuba, Nicolas Schilling, and Lars Schmidt-Thieme.
\newblock Learning data set similarities for hyperparameter optimization
  initializations.
\newblock In \emph{Metasel@ pkdd/ecml}, pages 15--26, 2015{\natexlab{a}}.

\bibitem[Feurer et~al.(2015{\natexlab{b}})Feurer, Springenberg, and
  Hutter]{feurer2015initializing}
Matthias Feurer, Jost Springenberg, and Frank Hutter.
\newblock Initializing bayesian hyperparameter optimization via meta-learning.
\newblock In \emph{Proceedings of the AAAI Conference on Artificial
  Intelligence}, volume~29, 2015{\natexlab{b}}.

\bibitem[Wistuba et~al.(2015{\natexlab{b}})Wistuba, Schilling, and
  Schmidt-Thieme]{wistuba2015learningHPOinits}
Martin Wistuba, Nicolas Schilling, and Lars Schmidt-Thieme.
\newblock Learning hyperparameter optimization initializations.
\newblock In \emph{2015 IEEE international conference on data science and
  advanced analytics (DSAA)}, pages 1--10. IEEE, 2015{\natexlab{b}}.

\bibitem[Wistuba et~al.(2015{\natexlab{c}})Wistuba, Schilling, and
  Schmidt-Thieme]{wistuba2015hyperparameter}
Martin Wistuba, Nicolas Schilling, and Lars Schmidt-Thieme.
\newblock Hyperparameter search space pruning--a new component for sequential
  model-based hyperparameter optimization.
\newblock In \emph{Joint European Conference on Machine Learning and Knowledge
  Discovery in Databases}, pages 104--119. Springer, 2015{\natexlab{c}}.

\bibitem[Perrone et~al.(2019)Perrone, Shen, Seeger, Archambeau, and
  Jenatton]{perrone2019learning}
Valerio Perrone, Huibin Shen, Matthias~W Seeger, Cedric Archambeau, and
  Rodolphe Jenatton.
\newblock Learning search spaces for bayesian optimization: Another view of
  hyperparameter transfer learning.
\newblock \emph{Advances in Neural Information Processing Systems},
  32:\penalty0 12771--12781, 2019.

\bibitem[Wistuba et~al.(2018)Wistuba, Schilling, and
  Schmidt-Thieme]{wistuba2018scalable}
Martin Wistuba, Nicolas Schilling, and Lars Schmidt-Thieme.
\newblock Scalable gaussian process-based transfer surrogates for
  hyperparameter optimization.
\newblock \emph{Machine Learning}, 107\penalty0 (1):\penalty0 43--78, 2018.

\bibitem[Pedregosa et~al.(2011)Pedregosa, Varoquaux, Gramfort, Michel, Thirion,
  Grisel, Blondel, Prettenhofer, Weiss, Dubourg, Vanderplas, Passos,
  Cournapeau, Brucher, Perrot, and Duchesnay]{scikit-learn}
F.~Pedregosa, G.~Varoquaux, A.~Gramfort, V.~Michel, B.~Thirion, O.~Grisel,
  M.~Blondel, P.~Prettenhofer, R.~Weiss, V.~Dubourg, J.~Vanderplas, A.~Passos,
  D.~Cournapeau, M.~Brucher, M.~Perrot, and E.~Duchesnay.
\newblock Scikit-learn: Machine learning in {P}ython.
\newblock \emph{Journal of Machine Learning Research}, 12:\penalty0 2825--2830,
  2011.

\bibitem[Ludwig et~al.(2020)Ludwig, Baracaldo, Thomas, Zhou, Anwar, Rajamoni,
  Ong, Radhakrishnan, Verma, Sinn, et~al.]{ludwig2020ibm}
Heiko Ludwig, Nathalie Baracaldo, Gegi Thomas, Yi~Zhou, Ali Anwar, Shashank
  Rajamoni, Yuya Ong, Jayaram Radhakrishnan, Ashish Verma, Mathieu Sinn, et~al.
\newblock {IBM Federated Learning}: an enterprise framework white paper v0. 1.
\newblock \emph{arXiv preprint arXiv:2007.10987}, 2020.
\newblock URL \url{https://github.com/IBM/federated-learning-lib}.

\bibitem[Feurer et~al.(2020)Feurer, Eggensperger, Falkner, Lindauer, and
  Hutter]{feurer-arxiv20a}
Matthias Feurer, Katharina Eggensperger, Stefan Falkner, Marius Lindauer, and
  Frank Hutter.
\newblock Auto-sklearn 2.0: The next generation.
\newblock In \emph{arXiv:2007.04074 [cs.LG]}, 2020.

\end{thebibliography}
}

\appendix
\newpage
\section{Appendix}

\subsection{Dataset details} \label{asec:emp:datasets}

The details of the binary classification data sets used in our evaluation is reported in Table~\ref{tab:datasets}. We report the 10-fold cross-validated balanced accuracy of the default HP configuration on each of data sets with centralized training. The ``Room for improvement'' column in Table~\ref{tab:datasets} denotes the difference between the best 10-fold cross-validated balanced accuracy obtained via centralized HPO and the 10-fold cross-validated balanced accuracy of the default HP configuration. 

\begin{table}[t]
\centering
\caption{OpenML binary classification data set details}
{\scriptsize
\begin{tabular}{llllcc}
\toprule
 Data & rows & columns & class sizes & Balanced accuracy on default & Room for improvement \\
\midrule
 EEG eye state & 14980 & 14 & (8257, 6723) & 90.28\% & 4.38\% \\
 Electricity & 45312 & 8 & (26075, 19237) & 87.78\% & 5.14\% \\
 Heart statlog & 270 & 13 & (150, 120) & 79.42\% & 6.17\% \\
 Oil spill & 937 & 49 & (896, 41) & 63.22\% & 11.36\% \\
 Pollen & 3848 & 5 & (1924, 1924) & 48.86\% & 3.35\% \\
 Sonar & 208 & 61 & (111, 97) & 87.43\% & 3.82\% \\
 PC3 & 1563 & 37 & (1403, 160) & 58.99\% & 4.82\% \\
\bottomrule
\end{tabular}
}
\label{tab:datasets}
\end{table}

\subsection{Search space} \label{asec:emp:search-space}

We use the search space definition used in the NeurIPS 2020 Black-box optimization challenge (\url{https://bbochallenge.com/}), described in details in the API documentation\footnote{\url{https://github.com/rdturnermtl/bbo_challenge_starter_kit/\#configuration-space}}. Given this format for defining the HPO search space, we utilize the following precise search space for the {\tt HistGradientBoostingClassifier} in {\tt scikit-learn}:
{\footnotesize
\begin{alltt}
api_config = \{
    'max_iter': {'type': 'int', 'space': 'linear', 'range': (10, 200)},
    'learning_rate': {'type': 'real', 'space': 'log', 'range': (1e-3, 1.0)},
    'min_samples_leaf': {'type': 'int', 'space': 'linear', 'range': (1, 40)},
    'l2_regularization': {'type': 'real', 'space': 'log', 'range': (1e-4, 1.0)},
\}
\end{alltt}
}

The HP configuration we consider for the single-shot baseline described in \S \ref{sec:emp} is as follows:
{\footnotesize
\begin{alltt}
api_config = \{
    'max_iter': 100,
    'learning_rate': 0.1,
    'min_samples_leaf': 20,
    'l2_regularization': 0,
\}
\end{alltt}
}

\end{document}